 \documentclass[pmlr,twocolumn,10pt]{jmlr} 

\usepackage{booktabs}
\usepackage{siunitx}

\usepackage{color}
\usepackage{xcolor, colortbl}
\usepackage{appendix}
\usepackage{psfrag}
\usepackage{booktabs,multirow,array,caption}
\usepackage{bm}
\usepackage{dsfont}
\usepackage{lscape}
\usepackage{hhline}
\usepackage{cite}
\usepackage{amsmath,amssymb,amsfonts}
\usepackage{algorithmic}
\usepackage{graphicx}
\usepackage{subcaption}
\usepackage{textcomp}
\usepackage{hyperref}
\usepackage{breakurl}
\usepackage{subcaption}

\newcommand{\bd}[1]{\boldsymbol{#1}}

\DeclareMathOperator*{\argmax}{argmax}

\theorembodyfont{\upshape}
\theoremheaderfont{\scshape}
\theorempostheader{:}
\theoremsep{\newline}

\jmlrvolume{LEAVE UNSET}
\jmlryear{2023}
\jmlrsubmitted{LEAVE UNSET}
\jmlrpublished{LEAVE UNSET}
\jmlrworkshop{Machine Learning for Health (ML4H) 2023} 

 \title[Time-dependent Probabilistic Generative Models for Disease Progression]{Time-dependent Probabilistic Generative Models for\\ Disease Progression}

\author{%
\Name{Onintze Zaballa}\Email{ozaballa@bcamath.org}\\
\addr BCAM-Basque Center for Applied Mathematics, Bilbao 48009, Spain
\AND
\Name{Aritz P\'erez} \Email{aperez@bcamath.org}\\
\addr BCAM-Basque Center for Applied Mathematics, Bilbao 48009, Spain
\AND
\Name{Elisa G\'omez-Inhiesto} \Email{elisa.gomezinhiesto@osakidetza.eus}\\
\addr Hospital Universitario Cruces, Barakaldo 48903, Spain
\AND
\Name{Teresa Acaiturri-Ayesta} \Email{mariateresa.acaiturriayesta@osakidetza.eus}\\
\addr Hospital Universitario Cruces, Barakaldo 48903, Spain
\AND
\Name{Jose A. Lozano} \Email{jlozano@bcamath.org}\\
\addr BCAM-Basque Center for Applied Mathematics, Bilbao 48009, Spain\\
\addr Intelligent Systems Group, University of the Basque Country UPV/EHU, Donostia 20018, Spain
}

\begin{document}

\maketitle

\begin{abstract}
Electronic health records (EHRs) contain valuable information for monitoring patients' health trajectories over time. Disease progression models have been developed to understand the underlying patterns and dynamics of diseases using these data as sequences. However, analyzing temporal data from EHRs is challenging due to the variability and irregularities present in medical records. We propose a Markovian generative model of treatments developed to (i) model the irregular time intervals between medical events; (ii) classify treatments into subtypes based on the patient sequence of medical events and the time intervals between them; and (iii) segment treatments into subsequences of disease progression patterns. We assume that sequences have an associated structure of latent variables: a latent class representing the different subtypes of treatments; and a set of latent stages indicating the phase of progression of the treatments. We use the Expectation-Maximization algorithm to learn the model, which is efficiently solved with a dynamic programming-based method. Various parametric models have been employed to model the time intervals between medical events during the learning process, including the geometric, exponential, and Weibull distributions. The results demonstrate the effectiveness of our model in recovering the underlying model from data and accurately modeling the irregular time intervals between medical actions.
\end{abstract}
\begin{keywords}
Disease Progression Modeling, Probabilistic Generative Model, Irregular Time, Electronic Health Records
\end{keywords}

\section{Introduction}
\label{sec:introduction}
Electronic health records (EHRs) contain a large amount of essential information for monitoring patients' health status throughout their clinical history. The temporal component of EHRs, which collects the sequence of medical events in the healthcare system over time, is important for understanding patients' treatment trajectories and identifying patterns in them. However, analyzing temporal data from medical records is a challenging task due to the variability and irregularities inherent in EHRs \citep{sarwar2022secondary}. Unlike common time-series data, where observations are recorded at regular intervals, EHRs possess irregular time intervals between patients' visits \citep{yadav2018mining}. 

Disease progression models have been developed to uncover the underlying patterns and dynamics of a disease \citep{zaballa2023learning, wang2014unsupervised, young2018uncovering}, and to predict medical outcomes from EHRs \citep{choi2016doctor, choi2016retain, alaa2019attentive}. Generative models, such as variations of Markov models, are commonly used in the literature to capture disease state transitions and model the temporal progression of diseases \citep{wang2014unsupervised,sukkar2012disease, severson2020personalized, ceritli2022mixture, huang2018probabilistic, liu2015efficient}. Although some of these approaches incorporate the concept of time into their models \citep{wang2014unsupervised, galagali2018patient, liu2015efficient}, they often focus on modeling the time intervals between hidden variables, rather than modeling the time elapsed between the observed ones, which is critical for estimating the time between consecutive medical events in real-world scenarios.

Deep learning techniques have been introduced to predict specific outcomes based on the progression of a disease \citep{choi2016doctor, choi2016retain, shickel2017deep}, with high prediction accuracy in future events but often overlooking the irregular temporality inherent in EHRs. While some methods have incorporated the irregular time information in their models \citep{shickel2017deep, choi2016doctor, pham2016deepcare, teng2020stocast, duan2019clinical}, they rarely focus on estimating the time intervals between consecutive medical events. Moreover, their lack of interpretability makes challenging the understanding of the underlying temporal evolution of a disease \citep{zhao2021event, shickel2017deep}. In contrast, probabilistic generative models enable the creation of representations of the temporal progression of a set of treatments through parametric modeling, providing physicians with more interpretable insights. These models, capable of being easily trained on unlabeled datasets with missing information, emerge as effective tools for modeling the underlying dynamics of a disease.

This paper presents an extension of the probabilistic generative model introduced in \citet{zaballa2023learning}. This method employs a latent class of treatments to categorize sequences of medical events into different subtypes and a latent sequence of stages to segment the sequence of events into subsequences of progression patterns. One of the key contributions of the present work is the incorporation of the time elapsed between medical actions within the sequence of events. With this approach, we aim to achieve the following objectives: (i) model the irregular time intervals between medical events; (ii) discover the different subtypes of disease progression in terms of the sequence of medical events and the time elapsed between them; and (iii) segment the sequences into progression patterns of treatments. 

The main contributions of this work are as follows:
\begin{itemize}
\setlength\itemsep{0em}
    \item We propose a probabilistic generative model based on Markov models that incorporates temporal information between medical events to model the underlying dynamics of disease treatments. Our model is flexible in terms of time distribution, allowing for the incorporation of the most appropriate distribution based on the available data. Specifically, we propose three parametric distributions to effectively model the irregular time intervals between medical actions: the geometric, exponential, and Weibull distributions.
    \item The model includes a class of treatments, which is a hidden variable that enables the grouping of patients based on the sequence of medical events and the time intervals between these events. Additionally, it incorporates a hidden sequence of progression stages, which segments treatments into distinct patterns of evolution. To efficiently learn the parameters of our generative model, we use the Expectation-Maximization algorithm, in which we propose a dynamic programming method for our specific model.
    \item We demonstrate the effectiveness of our approach in uncovering the underlying data model, predicting the irregular timing between medical events, and classifying treatments into different subtypes using synthetic and real-world data 
\end{itemize}
The remainder of this paper is organized as follows: Section 2 presents our proposed generative model and describes the methodology in detail. Section 3 presents the experimental setup and the results. Section 4 discusses the contributions and limitations of our approach. Finally, Section 5 draws the conclusions.

\section{Methodology}
\label{sec:methodology}

\subsection{Problem formulation}
\label{problem_formulation}

A patient's treatment associated with a disease, denoted by $\bd{a}$, is a sequence of medical events collected during repeated hospital visits. Let $A$ be the set of medical specialties (for instance, oncology, hematology, cardiology, etc), we define a patient's treatment as
$$\bd{a}=(a_1,...,a_m)$$
where $a_i \in A$ represents the $i$-th medical action of a patient. Each sequence of medical actions has an associated sequence of time intervals,
$$\bm{\tau}=(\tau_1,...,\tau_m)$$
where $\tau_i \in \mathbb{R}$ is the time interval between $a_{i-1}$ and $a_i$, for $i=2,...m$. We initialize $\tau_1$ as 0 to indicate the starting point of the treatment. 

Given a dataset of medical records, the objective is to develop a probabilistic generative model to effectively capture the temporal dynamics of the disease and the variability in treatment patterns.

\subsection{Model definition}
\label{model_definition}

We adopt the problem setting and notation as presented in \citet{zaballa2023learning}. This model is built on Markovian assumptions and considers that a sequence of actions has a structure of latent variables. These latent variables include the classes of treatments, which identify similar subtypes of treatments, and the stages, which segment each treatment into different progression patterns. We assume that all sequences of actions begin in the first stage, representing the initial steps of the treatment, and all classes of treatments have an equal number of stages. By defining these stages, it becomes possible to segment the sequences within each class of treatments into subsequences that are associated with their progression. Note that the same stage values from different classes of treatments represent different subsequences, which allows the model to be more flexible. Our primary contribution lies in expanding this model to include the irregular timing between consecutive medical actions, assuming that this timing varies depending on the latent class of treatment.

Let $\bd{a}=(a_1,...,a_m)$ be the treatment of a patient associated with a disease, where $a_i \in A$, and let $\bm{\tau}=(\tau_1,...,\tau_m)$ be the corresponding sequence of time intervals, where $\tau_i \in \mathbb{R}$. Let $\bd{s}=(s_1,...,s_m)$ be the sequence of latent stages associated with $\bd{a}$. The stages, denoted as $s_i$, belong to a set $S=\{1,...,r\}$ that represents all the possible stages of a treatment. Finally, let $c$ be the latent class of treatments which $\bd{a}$ belongs to. The class of treatments $c$ belongs to a set $C=\{1,...,k\}$ that represents all the possible classes, corresponding to distinct subtypes of treatments for a specific disease.

It is assumed that the progression stages are non-decreasing, implying that a sequence cannot go backward. Thus, for any given time point $i=1,...,m-1$, we have $s_i \leq s_{i+1}$. This assumption guarantees that the treatment moves forward without skipping any stage.

The proposal for the extended probabilistic generative model is as follows (see \figureref{fig:modeldescription}):

\begin{figure*}[htpb]
\centering
\includegraphics[width=10cm]{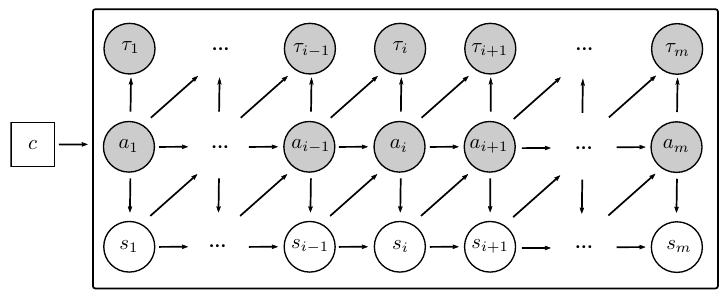}
\caption{Probabilistic generative model defined by the conditional distributions $p(a_i|a_{i-1}, s_{i-1},c)$, $p(s_i|a_{i}, s_{i-1},c )$ and $p(\tau_i|a_{i-1}, a_{i},c)$ for sequences of actions $\bd{a}$, sequences of time intervals $\bm{\tau}$, latent sequences of stages $\bd{s}$ and latent classes $c$. The gray figures represent the observed variables.}
\label{fig:modeldescription}
\end{figure*}

\begin{enumerate}
\setlength\itemsep{0em}
\item [a)] Draw a class of treatments $c\sim Mult(\bm{\theta}_C)$
\item [b)] Draw the initial medical action and the initial stage
$$a_1|c\sim~Cat(\bm{\pi}_A^c), \ \ s_1|a_1,c\sim~Cat(\bm{\pi}_S^{a_1,c}).$$
\item [c)] For each timestamp index $i$:
\begin{enumerate}
\setlength\itemsep{0em}
    \item [i)] Draw a medical action from $p(a_i|a_{i-1},s_{i-1},c)$, that is, \vspace{-0.2cm} $$a_{i}|a_{i-1},s_{i-1},c \sim Cat(\bm{\theta}^{a_{i-1},s_{i-1},c}_{A})$$
    \item [ii)] Draw a stage $s_i$ from $p(s_i|a_{i}, s_{i-1}, c)$,\vspace{-0.2cm} $$s_i|a_{i}, s_{i-1},c \sim Cat(\bm{\theta}_S^{a_{i},s_{i-1},c})$$
    \item [iii)] Draw the time interval from $p(\tau_i|a_{i-1}, a_i,c)$, that is, \vspace{-0.2cm} $$\tau_i|a_{i-1},a_i,c \sim F_T(\bm{\theta}_{T}^{a_{i-1},a_i,c})$$
\end{enumerate}
\end{enumerate}

Our generative model provides flexibility in capturing the time intervals between pairs of actions by utilizing an appropriate parametric distribution $F_T(\bm{\theta}_{T}^{a,a',c})$. It assumes that the time intervals depend on the latent class of treatments and pairs of actions, but not on the stage of progression. 

Translating the generative process into a joint probability model results in the expression:
\begin{align}
p(\bd{a}, \bm{\tau}, &\bd{s}, c) =&  \\
= & p(c) \prod_{i=1}^m p(a_i,s_i|a_{i-1}, s_{i-1},c) \cdot p(\tau_i | a_{i-1}, a_{i}, c), \nonumber \vspace{-0.5cm}
\end{align}
where
$$p(a_i,s_i|a_{i-1},s_{i-1},c)=p(a_{i}| a_{i-1}, s_{i-1},c)\cdot p (s_i | a_{i}, s_{i-1}, c)$$ 
and $p(a_1,s_1|a_0,s_0,c)=p(a_1,s_1|c)$. Furthermore, $s_1=1$, $a_{m}=end$, and $s_{i-1} \leq s_{i}$ for all $i$.

For each class, we define a Markov model to generate actions based on the previous action and stage in the sequence, and another Markov model to generate stages based on the previous stage and current action. These dependencies allow to maintain the consistency of the sequences of events over time. The distributions $F(\bm{\theta}_T)$ that we consider are the geometric, exponential and Weibull distributions.

The parameters of the initial model for medical actions and stages are denoted as
$\bm{\pi}_A^{c}$ and $\bm{\pi}_S^{a,c}$, respectively. Our goal is to estimate the parameters  $\bm{\theta} = \{ \bm{\theta}_C, \bm{\theta}_A, \bm{\theta}_S ,\bm{\theta}_T, \bm{\pi}_A, \bm{\pi}_S\}$ to capture the underlying dynamics and distributions in data.

\subsection{Maximum likelihood parameter estimation}
\label{subsec:MLE}

In this section, we introduce the procedure for learning the model parameters. Let $\set{D}=\{ (\bd{a}^i, \bm{\tau}^i) \}_{i=1}^N$ be the set of observed sequences of medical actions and time intervals, let $C$ be the set of latent classes of treatments and $S$ the set of latent stages of progression. We use the EM algorithm \citep{Bishop2006} to obtain the maximum likelihood estimate of the model's parameters in the presence of the latent variables. Due to the complete dataset is unavailable, we will instead consider the expected value of the log likelihood for the complete dataset under the posterior distribution of the latent variables, denoted as $p(c,\bd{s}|\mathbf{a},\bm{\tau})$. This involves considering all possible configurations for the hidden variables. Then, we solve the following maximization problem:
\begin{equation}
\label{eq:loglikelihood}
    \max_{\bm{\theta}} \sum_{(\bd{a},\bm{\tau}) \in D} \sum_{\bd{s}\in \set{S}_{\bd{a}}} \sum_{c \in C} p(\bd{s},c|\bd{a}, \bm{\tau}) \cdot \log  p(\bd{a}, \bm{\tau}, \bd{s}, c),
\end{equation}
where $\set{S}_{\bd{a}}$ is the set of all possible configurations of sequences of stages for $\bd{a}$. Note that every pair $(\bd{a}, \bm{\tau}) \in \set{D}$ contributes equally to the model regardless of its length due to
\begin{equation}
\label{eq:contribution}
\sum_{\substack{c\in C\\\bd{s}\in \set{S}_{\bd{a}} }} p(\bd{s},c|\bd{a}, \bm{\tau}) = \sum_{\substack{c\in C\\\bd{s}\in \set{S}_{\bd{a}} }} p(\bd{s}|c, \bd{a},\bm{\tau})\cdot p(c|\bd{a}, \bm{\tau}) = 1.
\end{equation}

The EM algorithm results in the following iterative process:

\noindent\textbf{E-step.} In this step, we calculate the posterior distribution of the latent variables given the observed data, that is, $p(\bd{s},c|\bd{a}, \bm{\tau})$. Then we use this posterior distribution to evaluate the expectation of the complete-data log-likelihood function as a function of the parameters $\bm{\theta}$ (Equation~\eqref{eq:loglikelihood}). The efficient learning procedure of these posterior distributions is performed with the dynamic programming-based method described in Appendix~\ref{appendix:dynamicprogramming}. This method is the adaptation of the conventional forward-backward algorithm used for HMMs to the characteristics of our generative model \citep{Bishop2006}.
 
\noindent\textbf{M-step.} In the maximization step, we maximize Equation~\eqref{eq:loglikelihood} using the posterior distributions computed in the E-step. This maximization is achieved using the Lagrange multiplier method. If $ \theta^{a,s,c}_{a' }, \theta^{a,s,c}_{s' }$ denote a component in $\bm{\theta}^{a,s,c}_A$,  $\bm{\theta}^{a,s,c}_S$, respectively, the model parameters corresponding to the transition from the pair $(a,s)$ to $(a',s')$ given the class $c$, where $a, a'\in A$ and $s,s' \in S$ are updated as follows:

\begin{align*}
   \theta^{a,s,c}_{a' }  =  & \dfrac{\sum_{ (\boldsymbol{a}, \bm{\tau}) \in\set{D}} \sum_{i=1}^{m_{\bd{a}}} \mathds{1}_{a,a'}(a_{i-1},a_i) \cdot p(s_{i} = s| c, \bd{a}, \bm{\tau}) }{\sum_{a \in A}  \sum_{(\boldsymbol{a}, \bm{\tau}) \in\set{D}} \sum_{i=1}^{m_{\bd{a}}} \mathds{1}_{a,a'}(a_{i-1},a_i) \cdot p(s_{i} = s| c, \bd{a}, \bm{\tau}) }\\
  \text{where} & \nonumber \\
   & \mathds{1}_{a,a'}(a_{i-1},a_{i})= 
    \begin{cases} 
      1             & \mbox{if } a_{i-1}=a, a_{i}=a'  \\
      0 & \mbox{otherwise.}
   \end{cases} \nonumber  
   & \nonumber \\
      & \nonumber \\
  \label{eq3:theta_states}
   \theta^{a,s,c}_{s' }= & \dfrac{\sum_{(\boldsymbol{a}, \bm{\tau}) \in\set{D}} \sum_{i=1}^{m_{\bd{a}}} \mathds{1}_{a}(a_i) \cdot p( s_{i-1}=s, s_i=s' | c,  \bd{a}, \bm{\tau}) }{\sum_{a\in A}  \sum_{(\boldsymbol{a}, \bm{\tau}) \in\set{D}} \sum_{i=1}^{m_{\bd{a}}} \mathds{1}_{a}(a_i) \cdot p(s_i=s' | c, \bd{a}, \bm{\tau}) }\\
   \text{where} & \nonumber \\
    &   \mathds{1}_{a}(a_{i})=  \nonumber
    \begin{cases} 
      1             & \mbox{if } a_{i}=a' \nonumber \\
      0 & \mbox{otherwise.}  \nonumber
   \end{cases} \nonumber 
  \end{align*}
  
If $\theta_{c}$ denotes a component in $\bm{\theta}_C$, the probability of the classes of treatments $c\in C$ is updated as

\begin{align*}
   \theta_{c} = & \dfrac{\sum_{(\boldsymbol{a}, \bm{\tau}) \in\set{D}} p(c|\bd{a}, \bm{\tau})}{\sum_{c\in C} \sum_{(\boldsymbol{a}, \bm{\tau}) \in\set{D}} p(c|\bd{a}, \bm{\tau})}
\end{align*}

As mentioned earlier, various distributions, such as geometric, exponential, or Weibull, can be used to model the time interval between each pair of actions within each latent class.
The parameters of the geometric distribution are updated as follows:
\begin{align*}
    \theta^{a,a',c}_T= \sum_{(\bd{a}, \bm{\tau}) \in \set{D}} \sum_{i=1}^{m_{\bd{a}}} \frac{n_{\bd{a}}}{\tau_i \cdot \mathds{1}_{a,a'}(a_{i-1},a_i) \cdot p(c|\bd{a}, \bm{\tau})+ n_{\bd{a}} } 
\end{align*}
where $n_{\bd{a}}= \mathds{1}_{a,a'}(a_{i-1},a_i) \cdot p(c|\bd{a}, \bm{\tau})$.

For the exponential distribution, which is the continuous analogue of the geometric distribution,
\begin{align}
        \theta^{a,a',c}_T  = \sum_{(\bd{a}, \bm{\tau}) \in \set{D}} \sum_{i=1}^{m_{\bd{a}}} \frac{n_{\bd{a}}}{\tau_i \cdot \mathds{1}_{a,a'}(a_{i-1},a_i) \cdot p(c|\bd{a}, \bm{\tau}) }  \nonumber
\end{align}

Finally, due to the absence of a closed-form solution for the maximum likelihood estimation of the Weibull distribution, it is necessary to employ numerical optimization methods to estimate the parameters (see \citet{lawless2011statistical} for more details). 

At each iteration of the algorithm, we combine the expectation and maximization steps for each $(\bd{a}, \bm{\tau}) \in \set{D}$ without the need to store the exponential number of probabilities for all configurations of sequences of stages and classes. Additionally, the proposed dynamic programming-based method (\appendixref{appendix:dynamicprogramming}) enables the EM algorithm to be solved while considering the exponential number of sequences of stages, with a computational complexity of $O(N\cdot m^2)$, where $m$ represents the length of the longest sequence of actions.

To simplify the notation and provide a clearer understanding of the model's main idea, we establish a fixed number of stages for all classes of treatments. Nevertheless, in scenarios where sequences remain incomplete due to ongoing treatments at the dataset's cutoff date, a more adaptable model can be formulated to accommodate varying numbers of stages. With this flexibility, the model can segment complete sequences into the maximum number of stages, denoted as $r^+$, while also handling incomplete sequences by using a reduced number of stages, ranging from $r^-$ to $r^{+}$.

\section{Experiments}
\label{sec:experiments}
In this section, we present the results obtained from a series of experiments conducted on both synthetic data and real-world data. Firstly, the experiments using synthetic data demonstrate the capability of our learning procedure to achieve a close approximation of the original generative model. Secondly, the experiments conducted on breast cancer patients show the applicability of the proposed model in gaining insights into the varying time intervals between consecutive medical records, as well as in the unsupervised classification of the treatments. 

\begin{figure*}[htpb]
\centering
\includegraphics[width=16cm]{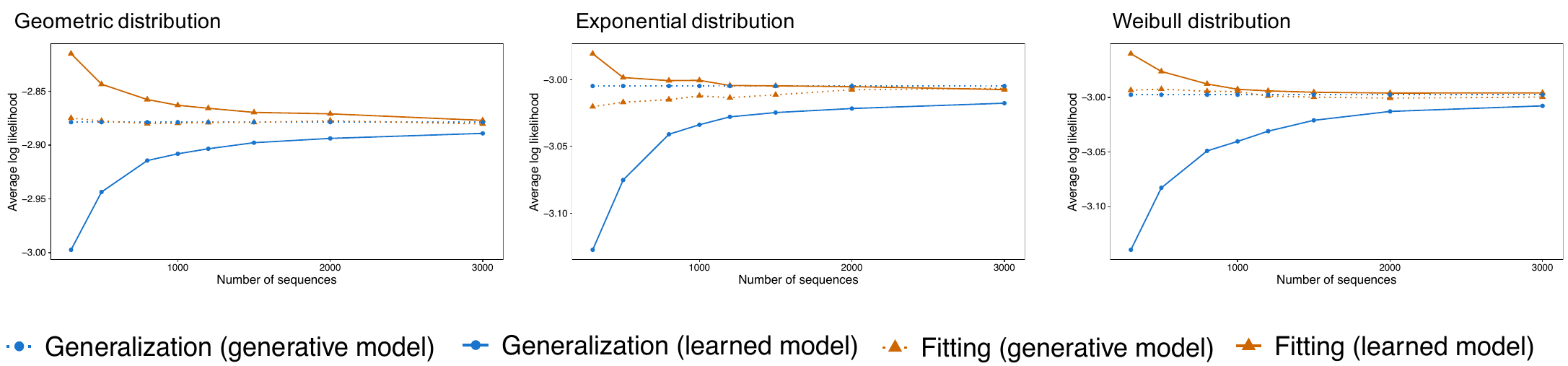}
\caption{Synthetic data results for different time distributions.}
\label{fig:synthetic}
\end{figure*}

\subsection{Synthetic data}
In this experiment, we demonstrate the learning performance of the proposed procedure concerning the number of training samples in practical scenarios. To do so, we use a set of artificially generated treatments derived from a randomly generated model. 

First, we create a probabilistic generative model $p_{\bm{\theta}}$, where the model's parameters are generated using the following procedure: $\bm{\theta}_C$ is sampled from a uniform Dirichlet distribution with parameters $\alpha=1$; similarly, $\bm{\theta}_A=\{\bm{\theta}^{a,s,c}_{A}\}$ is sampled from a uniform Dirichlet distribution with parameters $\alpha=1$ for each $a \in A$, $s\in S$ and $c \in C$; additionally, $\bm{\theta}_S= \{\theta^{s,a,c}_{S}\}$ is sampled from a Dirichlet distribution setting $\alpha=0.7$ for the parameters corresponding to transitions that remain in the same stage ($s'=s$) and setting $\alpha=0.3$ for the parameters related to transitions progressing to a different stage ($s'\neq s$), for each $a\in A$, $s,s'\in S$ and $c \in C$. The reason for setting a lower value when the transition progresses to a different stage is to generate more realistic sequences, avoiding excessively short subsequences of stages.

This experiment is repeated for each time distribution: geometric, exponential, and Weibull distributions. The parameters for the geometric distribution are sampled from a Beta(5,2) distribution, for the exponential distribution they are sampled from a Gamma(2,1) distribution, and for the Weibull distribution, the shape parameters are sampled from $\mathcal{U}(2,5)$, and the scale parameters are sampled from $\mathcal{U}(1,1.5)$.

For the sake of simplicity, we set a fixed total number of classes, $|C| = 2$, and define a range of stages from a minimum of $r^{-}=3$ to a maximum of $r^{+}=4$. These models allow us to generate training sets of various sizes, specifically $N=\{300,500, 800, 1000, 1200, 1500, 2000, 3000\}$, using the randomly generated model $p_{\bm{\theta}}$. We consider a set of $10$ unique actions to create these sequences. Additionally, we sample a test set of $4000$ sequences from $p_{\bm{\theta}}$ to evaluate the learning process.

To demonstrate that the learning algorithm can provide a good approximation of the original model with realistic training set sizes, we employ the EM-based procedure proposed in \sectionref{subsec:MLE} to fit the model on the training sets. For the EM initialization, we divide the observed sequences of actions into equal-length stage intervals. The initial parameters for the time distribution are uniform across all classes and are estimated with the observed time intervals between actions. For the initial class model, we initialize the probability of each sequence belonging to each class of treatments with the uniform distribution. We then add a probability $\epsilon=0.1$ to the true class to which they belong to prevent relabeling in the results. After learning the model, we analyze the evolution of the method's quality as the size of the training set, $n\in N$, increases. For each value of $n$, we obtain a new model $\bm{\theta}^n= \{\bm{\theta}_C^n, \bm{\theta}_A^n, \bm{\theta}_S^n, \bm{\theta}_{T}^n \}$ and assess its quality by computing the log likelihood of \equationref{eq:loglikelihood} normalized by $n$, making the datasets of different sizes comparable. 

The experiment is conducted five times for each time distribution, with each experiment considering a different random generative model, denoted as $p_{\bm{\theta}}$, from which the training sets and test sets are generated. \figureref{fig:synthetic} shows the fitting and generalization capabilities of our models by presenting the average log likelihood for the three time distributions. The solid orange line represents the average log likelihood of the learned models on the training sets, indicating how well the models fit the data. On the other hand, the solid blue lines represent the average log likelihood of the learned models on the test set, showing their ability to generalize to unseen data. The dotted lines correspond to the average log likelihood of the original generative models, with the orange line representing the training dataset and the blue line representing the test dataset. As we can see in \figureref{fig:synthetic}, as $n \in N$ increases, the curves representing the fitting and generalization of the learned models converge to the curves of the original generative models. This convergence indicates that, given a sufficiently large dataset, the proposed learning algorithm successfully recovers the original generative model that underlies the data.

\subsection{Real-world data}
\label{section:real_world_exp}
In this section, we show the utility of the generative model in real EHRs. We use our model in two different applications: for time interval prediction and for treatment classification.

\subsubsection{Dataset}
We use a dataset provided by the public healthcare system of the Basque Country, Spain. These EHRs consist of billing data and cover every outpatient and hospital visit of patients from 2016 to 2019. As a use case, we focus our attention on the breast cancer population, which comprises 645 patients. Their treatments average 115 medical actions, and they are generated by 23 unique medical specialties (selected following the procedure in \citet{zaballa2020identifying}). In total, there are 73150 transitions between pairs of actions, with a mean time interval of 10 days and a standard deviation of 31 days.

\subsubsection{Time prediction performance}
The goal of this experiment is to determine which parametric model provides better predictions for the time intervals between medical actions. To achieve this, our objective is to estimate the time interval until the next medical action as time progresses. 

\begin{table*}[htpb]
\caption{Mean absolute error in predicting the time interval until the next medical action.}
\centering
\begin{tabular}{r||r|r|r||r}
 & \multicolumn{3}{c||}{\textbf{Parametric}} & \textbf{Non-param.} \\
 & \textbf{Geometric} & \textbf{Exponential} &  \textbf{Weibull} & \multicolumn{1}{c}{\textbf{Median}} \\
\hline
\hline
\multicolumn{1}{r||}{\textbf{Empirical}}  & 16.36 & 17.06 & 18.03 & 3.86 \\
\hline
\multicolumn{1}{r||}{\citet{zaballa2023learning} \textbf{(mixture)}} & 4.64 & 4.62 &  4.24 & \\
\multicolumn{1}{r||}{\citet{zaballa2023learning} \textbf{(argmax)}} & 4.54 & 4.55 & 4.17 &  \\
\hline
\multicolumn{1}{r||}{\textbf{Proposed model (mixture)}} & 4.45 & 4.89 &  \textbf{4.12} & \\
\multicolumn{1}{r||}{\textbf{Proposed model (argmax)}} & 4.57 & 5.21 & 4.25 &  \\
\hline
\end{tabular}
\label{tab:MAE}
\end{table*}

\paragraph{Experiment Setup.}
We use a cross-validation approach to assess the predictive performance of the generative model. Following the results obtained in \citet{zaballa2023learning}, we consider 5 classes of treatments, with a minimum of 3 stages and a maximum of 4 stages for each treatment. In all training models, including the baselines, we use 90\% of the patients as the training set and 10\% as the test set. 

We train the models using the three time distributions: geometric, exponential, and Weibull. The initial parameters for the stages and time distributions are the same as in the synthetic experiments. However, for the initial class model, we use the K-medoids method \citep{zaballa2020identifying} for real-world data. Subsequently, we make predictions for each time step by sampling a set of time intervals from the learned generative model and using their median as the prediction for that time step. Let $\bd{a}_t=(a_1,\ldots,a_t)$ be the observed subsequence of actions up to time step $t$, and $\bm{\tau}_t=(\tau_1,\ldots,\tau_t)$ the observed subsequence of time intervals up to time step $t$. We define $q_t(c)$ as the probability distribution of classes given the subsequence of actions $\bd{a}_t$ and the subsequence of time intervals $\bm{\tau}_t$, in such a way that $q_t(c)$ changes as time progresses:
    \begin{equation*}
        q_t(c) = p(c|\bd{a}_t, \bm{\tau}_t).
    \end{equation*}
    
We estimate the time interval between medical actions, $\hat{\tau}_{t+1}$ for $t={2,\ldots,m}$, by sampling time intervals from the generative model in the following two ways:
\begin{enumerate}
\setlength\itemsep{0em}
    \item [(a)] Using the mixture of classes of treatments of the model, 
    \begin{equation}
    \vspace{-0.1cm}
    \sum_{c\in C} q_t(c)\cdot p(\tau_{t+1} | a_{t}, a_{t+1}, c) 
        \vspace{-0.1cm}
    \label{eq:mixture_prediction}
    \end{equation}
\item [(b)] Using the class of treatments of maximum probability, 
    \vspace{-0.1cm}
    \begin{equation}
     p(\tau_{t+1} | a_{t}, a_{t+1}, c^{*}), \ \ c^{*} = \argmax_{c \in C} q_t(c)
             \label{eq:argmax_prediction}
    \end{equation}
\end{enumerate}
The final prediction of the time interval $\hat{\tau}_{t+1}$ is given by the median of the samples obtained using (Equations \eqref{eq:mixture_prediction} and \eqref{eq:argmax_prediction}).

\paragraph{Evaluation metrics.} We evaluate the prediction error using the mean absolute error, that is, $|\tau - \hat{\tau}|$. 

\paragraph{Baselines.} On the one hand, we use parametric and non-parametric approaches to make predictions of the time interval until the next medical action. In the empirical parametric approaches, we fit the data to geometric, exponential and Weibull distributions, using $p(\tau|a,a')$ to estimate the time intervals. In the non-parametric approach, we predict the time using the median of the observed time intervals between each pair of medical actions. On the other hand, we compare our model against the one proposed by \citet{zaballa2023learning}. Since this model is not time-dependent, we first learn the generative model and then fit the geometric, exponential, and Weibull distributions to the training data as described in \sectionref{subsec:MLE}. We then use both the mixture of classes of the model (\equationref{eq:mixture_prediction}) and the class of maximum probability (\equationref{eq:argmax_prediction}) to sample time intervals and make the prediction with the median of these samples.

\begin{figure*}[htpb]
\centering
\includegraphics[width=16cm]{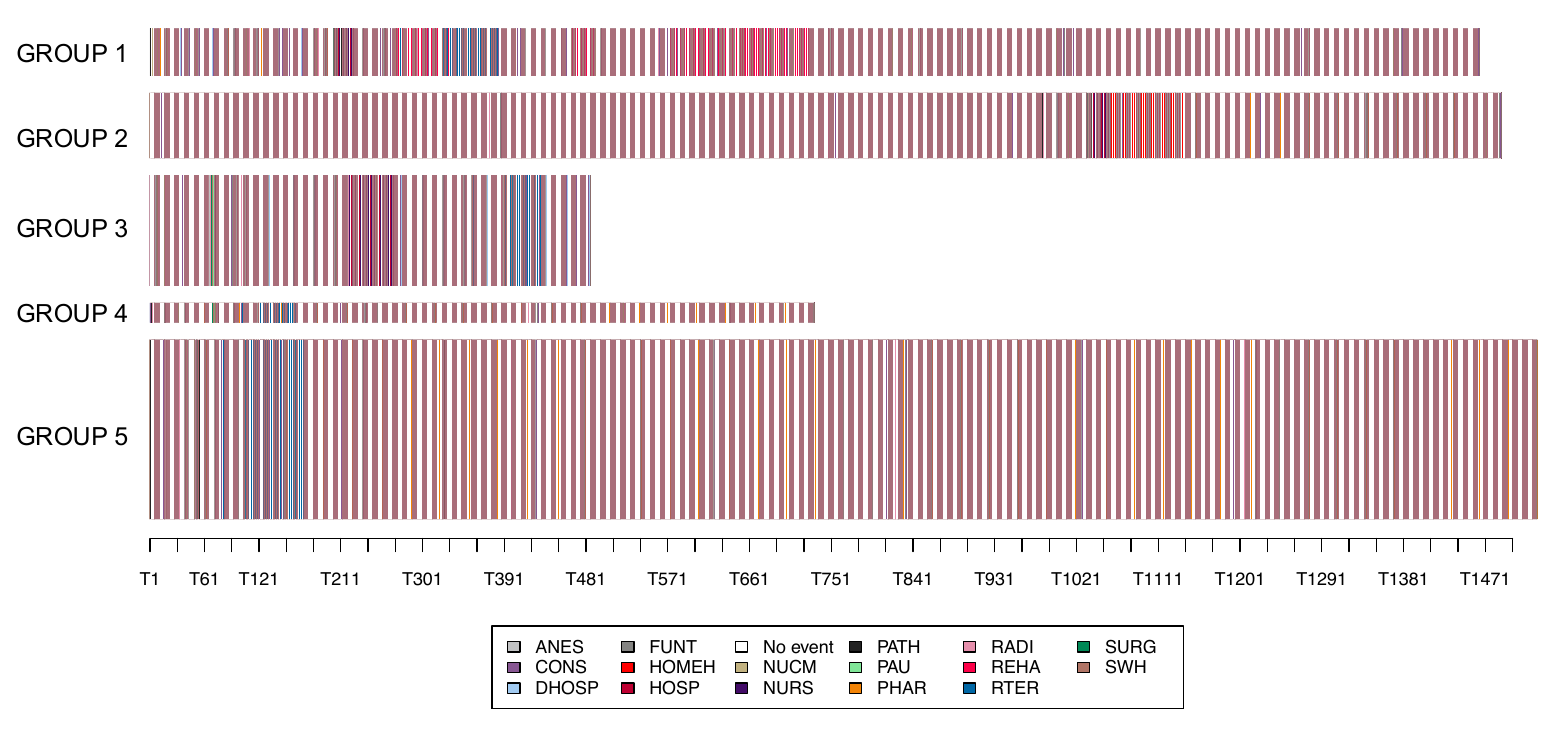}
\caption{Classification results for treatments associated with breast cancer considering the time between medical events. See \appendixref{appendix:medical_actions_legend} for the description of the medical actions, and see \appendixref{app:treatment_classification} for the representation without displaying the time intervals (\textit{No event}).}
\label{fig:weibull_rep_time}
\end{figure*}

\paragraph{Prediction performance.} \tableref{tab:MAE} compares the results from various algorithms, confirming that our proposed approach outperforms baseline models in the parametric setting. Specifically, predictions using the Weibull distribution show the lowest mean absolute error among these models. For more details on errors made by different approaches when predicting the most frequent pairs of actions, refer to \appendixref{app:heatmaps}. We can conclude that the Weibull distribution performs better than other parametric approaches, and that fitting the time in the learning process enhances prediction accuracy. It is important to mention that the superior performance of the non-parametric method can be due to the robustness of the median when handling extreme time interval values that deviate from the mean. The proposed parametric approaches, however, are more sensitive to these time values.

The more accurate predictive performance of the non-parametric method in Table~\ref{tab:MAE} can be attributed to the robustness of the median when handling extreme time interval values that deviate significantly from the mean.

\subsubsection{Treatment classification}
In this second experiment, we aim to explore the impact of incorporating time modeling on the representation of treatment subtypes. Using the same hyperparameters as in the previous section, we trained the model using the EM-based procedure described in \sectionref{subsec:MLE}. The classification of treatments is carried out by associating each sequence of actions $\bd{a}$ and its corresponding $\bm{\tau}$ with the most probable class $c^{*}$, that is, 
\begin{equation}
\label{eq:maxclass}
    c^{*}= \argmax_{c \in C} p(c|\bd{a}, \bm{\tau}).
\end{equation}
The dynamics of the sequences of actions of each class are characterized by a representative sequence. This is defined as the most probable pair $(\bd{a}, \bm{\tau})$ within each class normalized by the length of $\bd{a}$, in order to avoid the probability $p(\bd{a}, \bm{\tau}|c)$ to exponentially decrease as long as the length of $\bd{a}$ increases. That is, 
\begin{equation}
\label{eq:maxaction}
    \bd{a}^{*}= \argmax_{\bd{a}} \frac{\log p(\bd{a}, \bm{\tau}|c)}{|\bd{a}|}.
\end{equation}

\begin{figure}[h!]
\centering
\includegraphics[width=\linewidth]{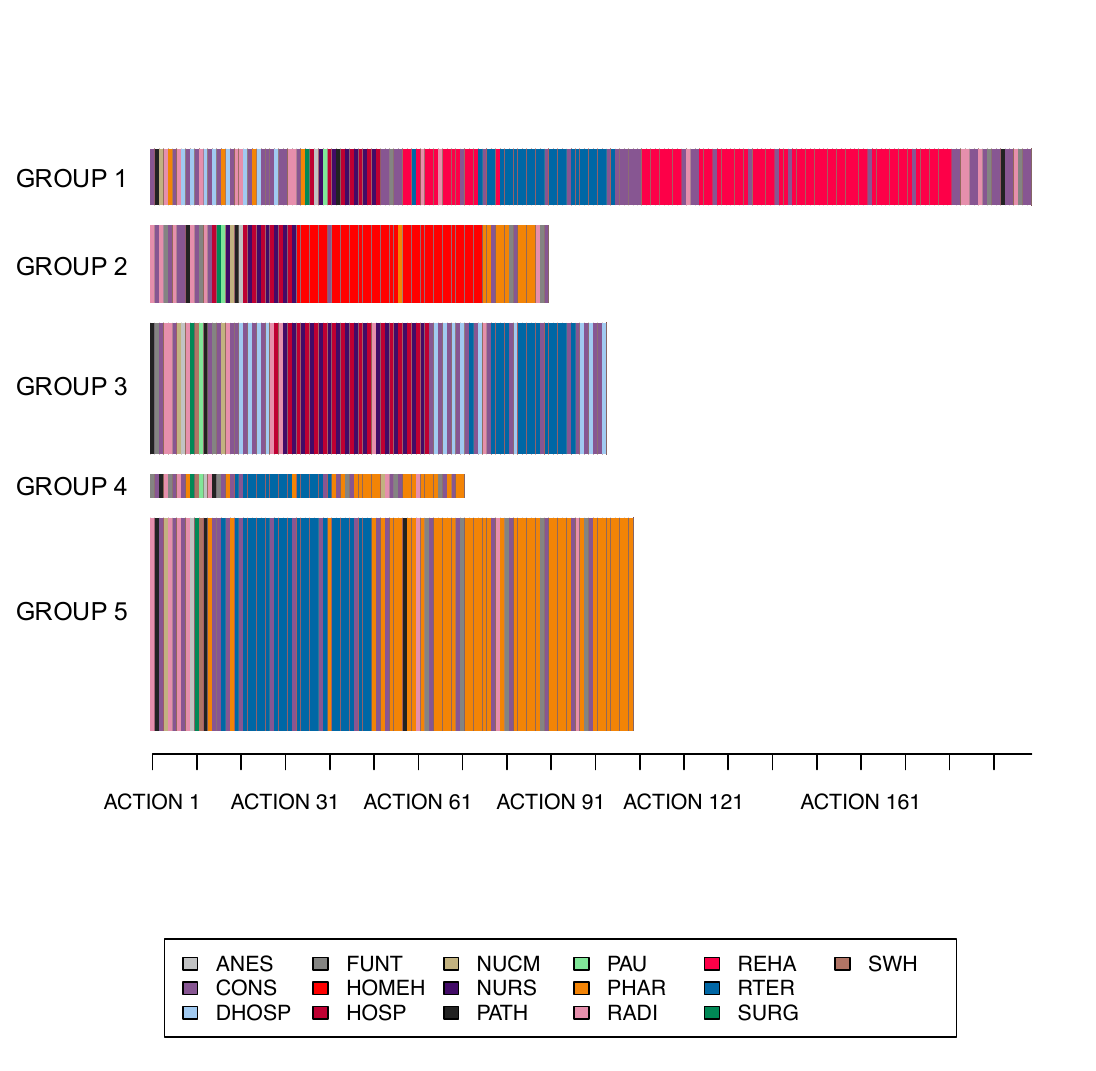}
\caption{Classification results for treatments associated with breast cancer without representing the time intervals between the medical actions.}
\label{fig:weibull_rep_notimeint}
\end{figure}

\figureref{fig:weibull_rep_time} presents the five representative breast cancer treatments obtained using the Weibull distribution, which is the distribution with the best results in the previous experiment. These treatments characterize different progression subtypes. Figure~\ref{fig:weibull_rep_notimeint} offers a more interpretable view of these results in terms of treatment patterns, displaying the same outcomes as Figure~\ref{fig:weibull_rep_time} but without showing the time intervals.

The major patterns of the representative treatments, which consists of real sequences of medical actions from EHRs, are as follows:
\begin{itemize}
\setlength\itemsep{0em}
    \item \textbf{Group 1. } Chemotherapy + Surgery + Hospitalization + Radiotherapy + Rehabilitation (11.3 \%)
    \item \textbf{Group 2. } Surgery + Hospitalization + Home hospitalization + Hormonotherapy (18.2 \%)
    \item \textbf{Group 3. } Surgery + Chemotherapy + Hospitalization + Radiotherapy (24\%)
    \item \textbf{Group 4. } Surgery + Radiotherapy + Hormonotherapy (5\%)
    \item \textbf{Group 5. } Surgery + Radiotherapy + Hormonotherapy (41.5\%)
\end{itemize}

Figure~\ref{fig:weibull_rep_time} shows that all the treatments start with the diagnosis process (conducted through radiology, nuclear medicine and pathological anatomy medical services). After receiving the specific therapy for each group, patients undergo regular follow-up consultations and medical tests. Note that Group 4 and Group 5 seem to be similar subtypes, however, their primary distinction lies in the longer duration of the treatment for patients in Group 5. All these findings related to the treatment patterns and their duration align with clinical practice guidelines \citep{ESMO1}.

\appendixref{app:treatment_classification} shows the comparison of these results with the representative treatments identified using the model in \citet{zaballa2023learning}.

\section{Discussion}
\label{sec:discussion}
This work proposes a probabilistic generative model that incorporates temporal information between medical events to model the underlying dynamics of disease treatments. This model is flexible in terms of time distribution, enabling the adoption of the most suitable distribution for the available data. Specifically, we propose three parametric distributions to effectively model the irregular time intervals between medical actions: the geometric, exponential, and Weibull distributions. The model includes a latent class variable, which makes the time modeling a mixture of these parametric distributions.

Unlike existing disease progression models \citep{shickel2017deep, choi2016doctor, teng2020stocast, wang2014unsupervised, galagali2018patient}, this is the first generative model of sequences that primarily aims to comprehend the temporal evolution of a disease, taking into account the temporal irregularities between observed medical events. Our approach provides interpretable representations of the temporal progression within sequences of actions through parametric modeling, by simultaneously capturing both disease stage transitions and distinct disease subtypes. We would like to emphasize that the main focus of this model is on learning the underlying distribution of a set of sequences of medical events. By capturing the temporal dynamics of these sequences, we open up a wide range of potential applications, including the prediction of medical variables, treatment classification, and the generation of new treatments, as demonstrated in our experiments.

The proposed model significantly outperforms the parametric baselines in predicting time intervals between medical events, as shown in Table~\ref{tab:MAE}. These results highlight the importance of considering treatment classes and progression stages for modeling the irregular time gaps within sequences of actions. The second set of experiments uses a modification of the model presented in \citet{zaballa2023learning}. This model originally does not consider time information, however, to be able to compare our model with a baseline, we introduced time interval estimation after the original model was already learned. Note that the structure of both models is similar in terms of classes and stages, which may explain their similar predictive results. However, our proposed model is able to slightly improve the predictive results by jointly learning the time intervals and latent variables, and provides a more informative representation of data in terms of medical actions and their time intervals.

The more accurate predictive performance of the non-parametric method in Table~\ref{tab:MAE} can be attributed to the robustness of the median when handling extreme time interval values that deviate significantly from the mean. Our proposed parametric probability distributions are more sensitive to these outliers and may not adequately approximate to these extreme time intervals. Nevertheless, the difference in the mean absolute error of the non-parametric method and our model is just 0.26 days. 


Finally, although parametric models, in particular the Weibull distribution, have shown favorable results for time estimation, they may not capture the full complexity and variability present in data. In future work, we aim to address this limitation by incorporating non-parametric techniques into our approach. For instance, non-parametric kernel density estimation \citep{malec2014nonparametric} could provide even more flexibility to the model and potentially capture a wider range of patterns and distributions in the time intervals.

\section{Conclusion}
\label{sec:conclusion}

In conclusion, this work presents a comprehensive framework for incorporating temporality into disease progression modeling. The main contribution is the proposal of a time-dependent probabilistic generative model for unsupervised classification of treatments with irregular time intervals. The generative model allows to: (i) model the irregular time intervals between medical events; (ii) discover the different subtypes of disease progression in terms of the sequence of medical events and the time elapsed between them; and (iii) segment the sequences into progression patterns of treatments. 

We validate this approach through a simulation experiment, successfully recovering the original model. Additionally, we demonstrate, using real EHRs, that the model accurately captures underlying temporal dynamics and variability within treatment subtypes. Practical applications of this model include assessing the adherence of treatment trajectories to medical practice guidelines, simulating new treatments, predicting the timing of the next hospital visit, and generating an interpretable data taxonomy for a comprehensive understanding of the disease.

\acks{This work has been supported by the Basque Government through the BERC 2022-2025 program and BMTF project, and by the Ministry of Science, Innovation and Universities: BCAM Severo Ochoa accreditation CEX2021-001142-S/MICIN/AEI/10.13039/501100011033. Jose A. Lozano is also supported by the Basque Government under grant IT1504-22 and Ministry of Science and Innovation under grant PID2022-137442NB-I00. Onintze Zaballa also holds a predoctoral grant EJ-GV 2019 from the Basque Government.}

\bibliography{jmlr-sample}

\newpage
\appendix

\clearpage

\onecolumn
\section{Efficient inference based on dynamic programming}
\label{appendix:dynamicprogramming}

Training a generative model poses a significant challenge, especially when dealing with large datasets and long sequences. Exact parameter learning for the model can be computationally expensive in such cases. To address this, we use a similar strategy as in the forward-backward algorithm, which is a dynamic programming-based method used to compute the posterior marginal distribution of hidden states in Hidden Markov Models \citep{Bishop2006}. This method reduces the number of computations required, thus improving the overall efficiency of our approach. This inference plays an important role in the learning process of the model, particularly in the E-step, where we need to find the posterior distribution of the latent variables $p(\bd{s},c|\bd{a},\bm{\tau})$. We then use this posterior distribution to evaluate the expectation of the logarithm of the complete-data likelihood function \equationref{eq:loglikelihood_app}, as a function of the parameters $\bm{\theta}~=~\{ \bm{\theta}_C, \bm{\theta}_A, \bm{\theta}_S, \bm{\theta}_{T} \}$.

\begin{equation}
\label{eq:loglikelihood_app}
    \max_{\bm{\theta}} \sum_{(\bd{a},\bm{\tau}) \in D} \sum_{\bd{s}\in \set{S}_{\bd{a}}} \sum_{c \in C} p(\bd{s},c|\bd{a}, \bm{\tau} ) \cdot \log  p(\bd{a}, \bm{\tau}, \bd{s}, c)
\end{equation}
where $\set{S}_{\bd{a}}$ is the set of all the compatible sequences of stages for $\bd{a}$, and $\bm{\theta}~=~\{\bm{\theta}_C, \bm{\theta}_A, \bm{\theta}_S, \bm{\theta}_{T}, \bm{\pi}_A, \bm{\pi}_S\}$. 

Let us assume that we have a training set $\set{D}=\{ (\bd{a}^i, \bm{\tau}^i) \}_{i=1}^N$ that consists of a set of treatments $\bd{a}=(a_1,....,a_m)$ and their corresponding sequences of time intervals $\bm{\tau}=(\tau_1,...,\tau_m)$. Let consider the underlying sequence of latent stages $\bd{s}=(s_1,...,s_m)$ where $s_i\in S$, and a latent variable of classes $c\in C$ for each pair $(\bd{a}, \bm{\tau})\in \set{D}$. We aim to estimate the maximum likelihood parameters $\bm{\theta}$ of the model in each iteration of the EM algorithm. 

We are interested in finding the posterior distribution $p(s'| \bd{a}, \bd{\tau}, c)$ and $p(s, s'| \bd{a}, \bd{\tau}, c)$ for $s',s \in S$ to learn the maximum likelihood estimate parameters. For this, we need to marginalize $p(\bd{s}|\bd{a}, \bd{\tau}, c)$ and compute the probability of each possible stage $s$ at time $t$ in each possible class $c$. That is, the probability of all the sequences of stages with the form $(s_1,...s_{t-1}, s , s_{t+1},...,s_{m})$ in $c$. Recall that this requires $\binom{m-2}{r-1}$ number of configurations for $\bd{s}$ (the last stage is fixed), which is exponential. 

Let us assume that $f_c(i,s)$ is the sum of the probabilities of all the sequences of stages $(s_1,...,s_i)$ in the class $c$ that end at $s_i=s$, and $g_c(i,s)$ is the sum of the probabilities of all the sequences of stages $(s_{i+1},...,s_m)$ that start at $s_i=s$ in the class $c$.
Then,
\begin{align}
f_c(i,s)&= \sum_{\bd{s}_{1:i}} p(\bd{s}_{1:i},\bd{a}_{1:i},\bm{\tau}_{1:i}| c) \label{f_t} \\
g_c(i,s)&= \sum_{\bd{s}_{i+1:m}} p(\bd{s}_{i+1:m}, \bd{a}_{i+1:m},\bm{\tau}_{i+1:m}| s_{i}= s,c), \label{g_t}
\end{align}
where $\bd{a}_{j:k}=(a_j,....,a_k)$, $\bm{\tau}_{j:k} = (\tau_j,...,\tau_k)$ and $\bd{s}_{j:k}=(s_j,....,s_k)$.

Now, we can express the sum of the probabilities of the sequences for which $\bd{s}_{i-1,i}=(s,s')$ as 

\begin{align*}
p(s_{i-1} = s, s_{i} = s'| \bd{a}, \bm{\tau} ,c )& = \frac{p(s_{i-1}=s,s_i=s',\bd{a}, \bm{\tau}|c)}{p(\bd{a}, \bm{\tau} |c)}
\end{align*}

Using Equations~\eqref{f_t} and \eqref{g_t},

\begin{align*}
p(s_{t-1}=s,s_t=s', \bd{a}, \bm{\tau} |c) & = \sum_{\substack{\bd{s}_{1:i-2}\\ \bd{s}_{i+1:m}}} p(\bd{s}_{1:i-2},s_{i-1}=s,\bd{a}_{1:i-1}, \bm{\tau}_{1:i-1}|c) \cdot p(a_i|a_{i-1},s_{i-1}= s,c)\cdot  \\
& \hspace{1.5cm}  p( s_i= s' |  a_{i},s_{i-1}=s, c) \cdot p(\tau_i|a_{i-1},a_{i}, c) \cdot p(\bd{s}_{i+1:m},\bd{a}_{i:m}, \bm{\tau}_{i:m}| s_i=s',c)\\
& = f_c(i-1,s)\cdot p(a_i|a_{i-1},s_{i-1}=s,c)\cdot  p(s_{i}=s'|a_i,s_{i-1}=s,c) \cdot p(\tau_i|a_{i-1},a_{i},c) \cdot g_c(i,s')  \nonumber \\ 
\end{align*}

We can store the values obtained from the functions $f_c$ and $g_c$ for $t\in \{1,...,m\}$ and $s\in S$ in a matrix of size $r\times m$ associated with each function. Using dynamic programming, we efficiently compute $f_c$ and $g_c$ and reduce the number of computations for the parameter estimation. The functions $f_c$ and $g_c$ are defined as recursive functions as follows

\begin{align*}
f_c(i,s) =  & f_c(i-1,s)\cdot p(a_i|a_{i-1}, s-1,c)  \cdot p(s|a_{i},s,c)\cdot  p(\tau_{i}|a_{i-1}, a_i,c) \\ 
& + f_c(i-1,s-1) \cdot p(a_i|a_{i-1}, s-1,c) \cdot p(s|a_{i}, s-1,c) \cdot p(\tau_{i}|a_{i-1}, a_i,c) \\
g_c(i,s)= & g_c(i+1,s+1) \cdot p(a_{i+1}|a_{i},s+1,c) \cdot p(s+1|a_{i+1}, s,c) \cdot p(\tau_{i+1}|a_{i},a_{i+1},c)\\
& + g_c(i+1,s) \cdot   p(a_{i+1}|a_{i},s,c) \cdot p(s|a_{i+1}, s,c)\cdot p(\tau_{i+1}|a_{i},a_{i+1},c)
\end{align*}

The functions $f_c$ and $g_c$ are defined in such a way that consecutive stages are non-decreasing. The dynamic programming method significantly reduces the number of computations for the parameter estimation. Intuitively, instead of individually computing the posterior distributions of the latent variables, $p(\bd{s},c|\bd{a}, \bm{\tau})$ for each possible configuration of $\bd{s}$ and $c$, dynamic programming reuses the transition probabilities that sequences share to reduce the number of computations. 

Finally, to model the time variable, we use the cumulative distribution function $F(x;\bm{\theta})$ for the exponential and Weibull distributions, given their continuous nature. In these cases, $p(\tau | a, a', c)$ is computed as $1-F(\tau;\bm{\theta}_T)$. However, for the geometric distribution, we use the probability density function, that is, $p(\tau | a, a', c) = f(\tau;\bm{\theta}_T)$.

\clearpage
\section{Time prediciton error in real EHRs}
\label{app:heatmaps}
This appendix presents the mean absolute errors for time interval predictions in Figures \ref{heatmap_empirical} and \ref{heatmap_model}. These mean absolute errors are calculated for the most frequent transitions between medical specialties, allowing us to demonstrate the improvements in predictions made by our model compared to empirical parametric methods.

\begin{figure}[ht!]
  \floatconts
  {heatmap_empirical}
  {\caption{Heatmap of the mean absolute errors of the prediction of time intervals using the empirical distributions.}}
  {%
    \subfigure[Geometric distribution (empirical)]{\label{fig:geom_real}%
      \includegraphics[width=0.3\linewidth]{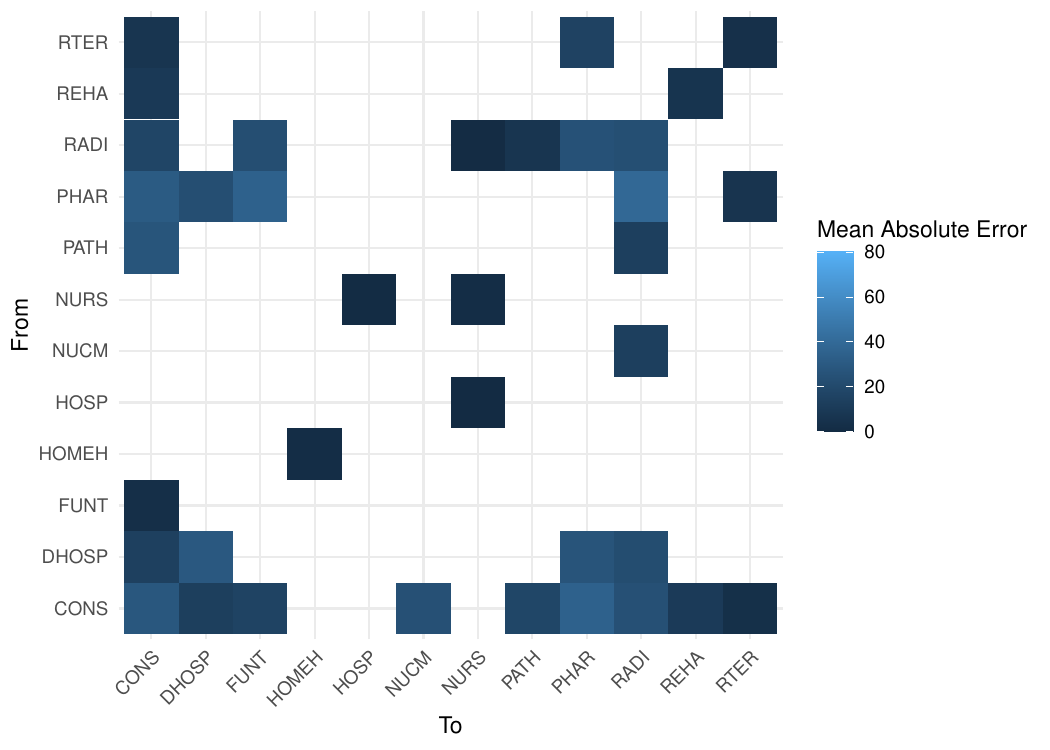}}%
    \qquad
    \subfigure[Exponential distribution (empirical)]{\label{fig:exp_real}%
      \includegraphics[width=0.3\linewidth]{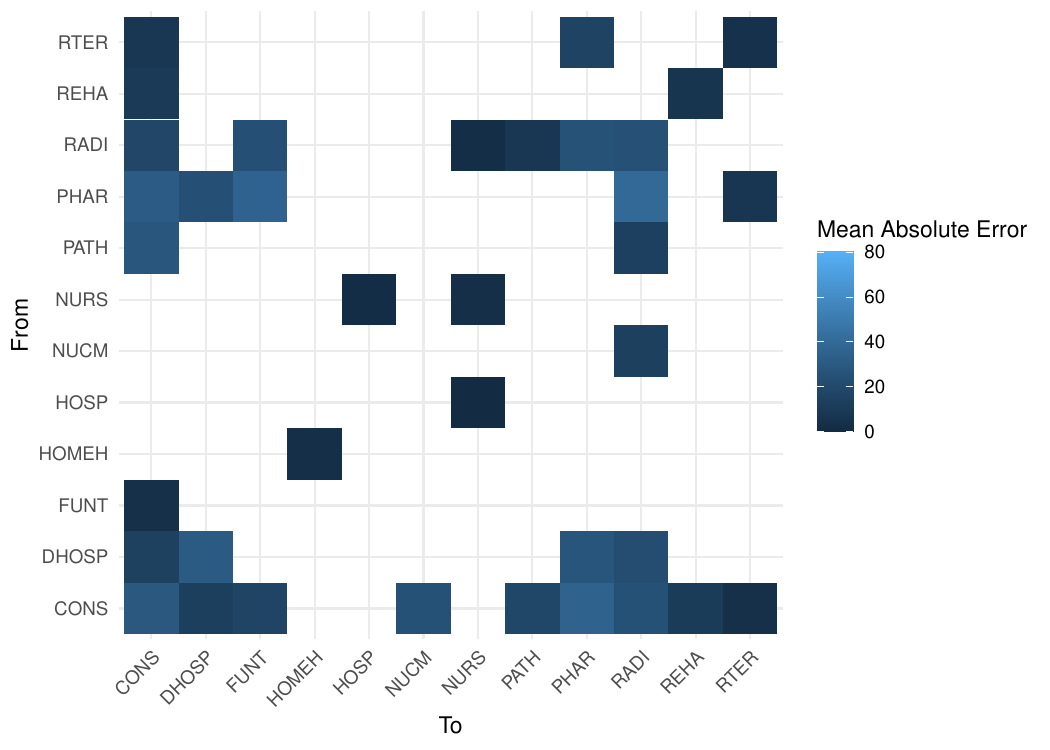}}
    \qquad
    \subfigure[Weibull distribution (empirical)]{\label{fig:weibull_real}%
      \includegraphics[width=0.3\linewidth]{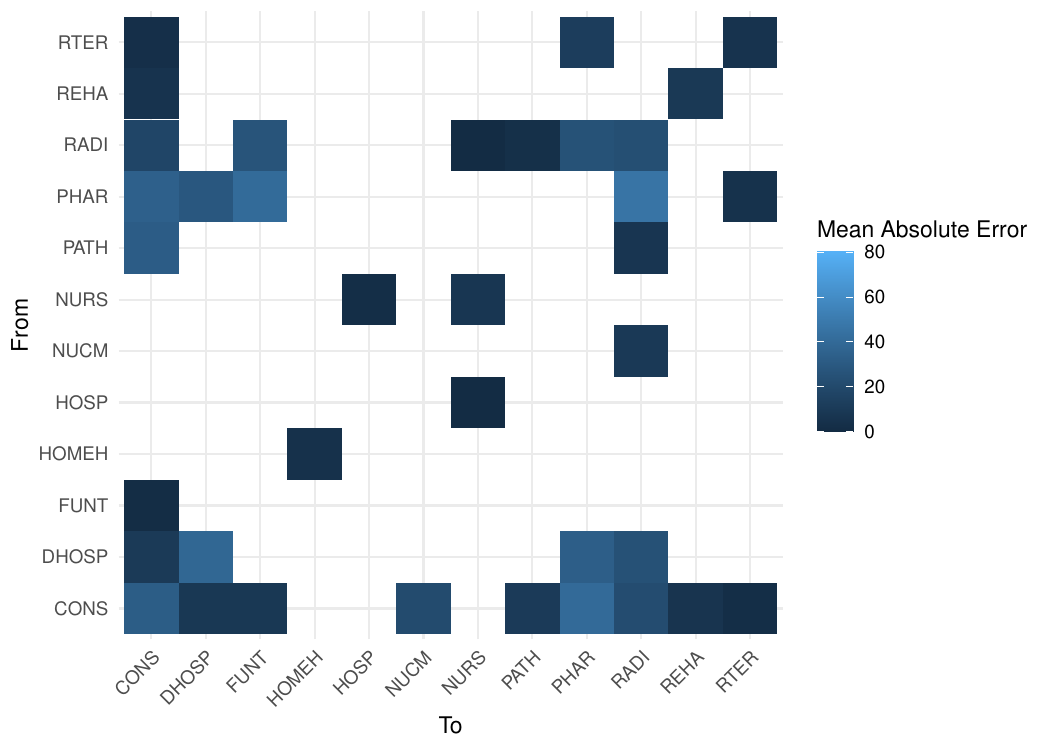}}
    \qquad
  }
\end{figure}

\begin{figure}[ht!]
  \floatconts
  {heatmap_model}
  {\caption{Heatmap of the mean absolute errors of the prediction of time intervals using the proposed generative model. The results of our model are obtained from the mixture of classes \equationref{eq:mixture_prediction}.}}
  {%
    \subfigure[Geometric distribution (proposed model)]{\label{fig:geom_mixture}%
      \includegraphics[width=0.3\linewidth]{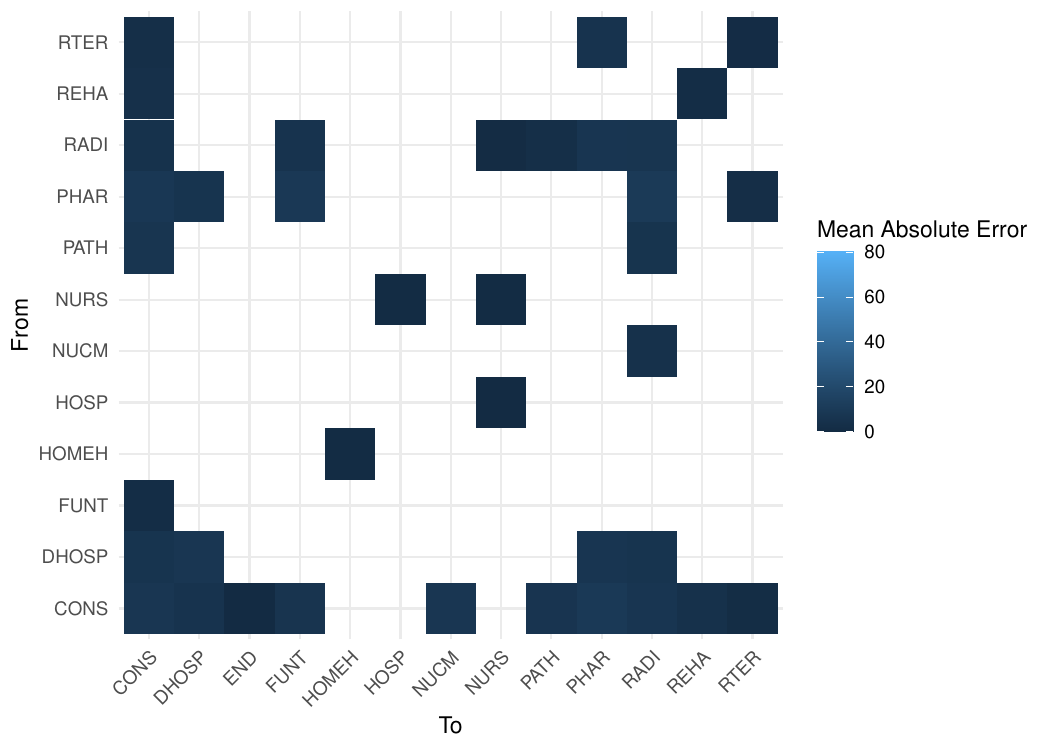}}%
    \qquad
    \subfigure[Exponential distribution (proposed model)]{\label{fig:exp_mixture}%
      \includegraphics[width=0.3\linewidth]{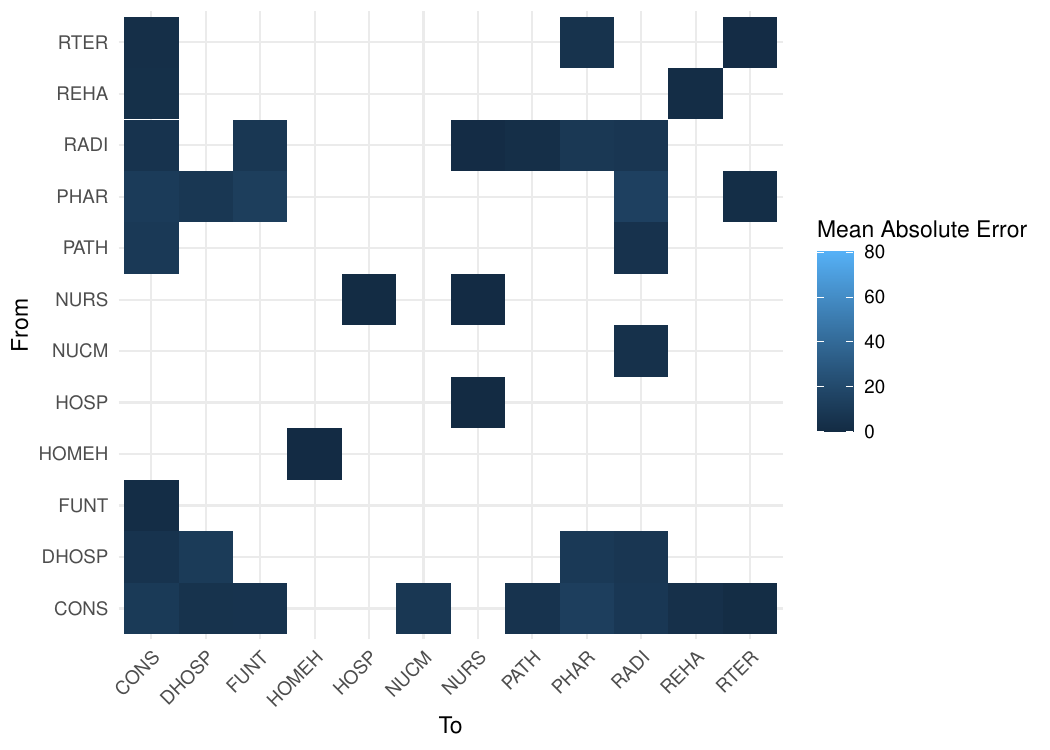}}
    \qquad
    \subfigure[Weibull distribution (proposed model)]{\label{fig:weibull_mixture}%
      \includegraphics[width=0.3\linewidth]{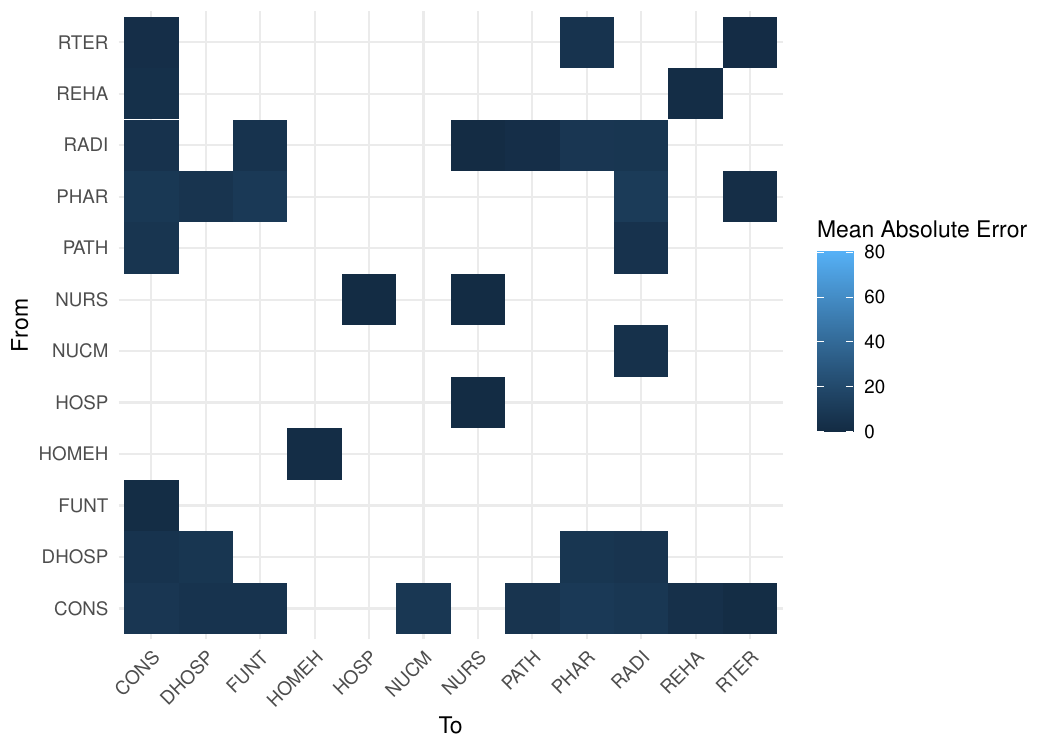}}
  }
\end{figure}

\clearpage
\section{Description of the medical actions of real EHRs.}
\label{appendix:medical_actions_legend}

In this appendix we explain the abbreviation and description of each medical action of the real breast cancer data in \sectionref{section:real_world_exp}.

\begin{table}[ht!]
\centering
\caption{Description of the medical actions}
\begin{tabular}{ l l } 
\hline
\textbf{Abbreviated form} & \textbf{Full Form}\\
 \hline 
 ANES & Anesthesia  \\ 
 CONS & Consultation  \\  
 DHOSP& Day Hospital \\  
 FUNT & Functional Testing \\  
 HOMEH & Home Hospitalization \\ 
 HOSP & Hospitalization \\ 
 NUCM & Nuclear Medicine \\ 
 NURS & Nursing Unit \\ 
 PATH & Pathological Anatomy \\  
 PAU & Post Anesthesia Care Unit \\  
 PHAR & Pharmacy \\ 
 RADI & Radiology \\  
 REHA & Rehabilitation \\ 
 RTER & Radiotherapy \\
 SURG & Surgery Unit \\ 
 SWH & Surgery without Hospitalization \\ 
 \hline
\end{tabular}
\label{action_description}
\end{table}

\clearpage
\section{Comparison of treatment classification outcomes with real-world data}
\label{app:treatment_classification}
In this appendix, we compare the representative treatment outcomes obtained with the proposed time-dependent generative model with the results obtained obtained with the model developed in \citet{zaballa2023learning}, which does not account for the temporal component. In both cases, horizontal lines represent treatment subtypes, vertical lines represent medical actions, and the width of the representative sequences is proportional to the number of patients in each subtype. 

\begin{figure}[t]
\centering
\includegraphics[width=0.6\linewidth]{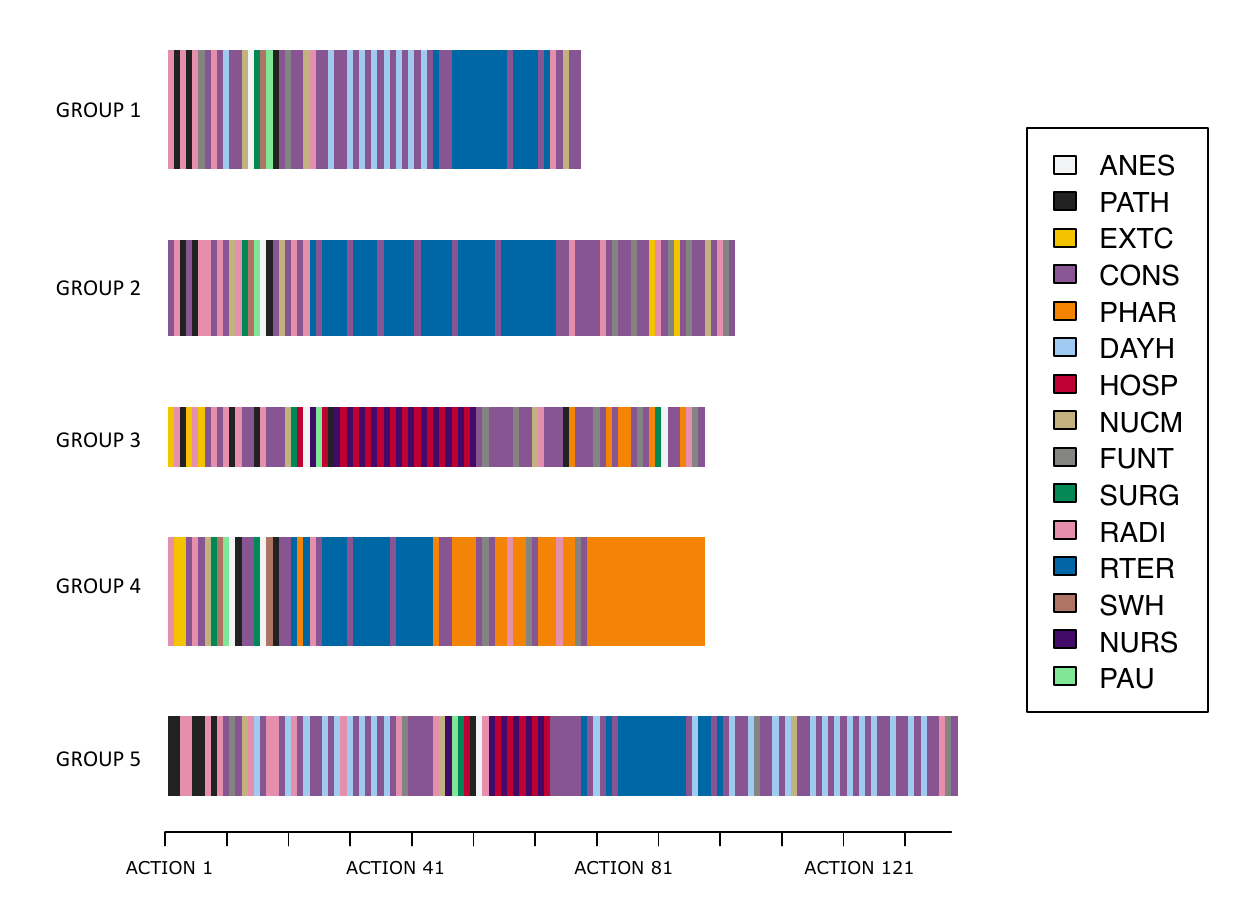}
\caption{Classification results for treatments associated with breast cancer obtained with the model in \citet{zaballa2023learning}.}
\label{fig:dp_notime}
\end{figure}

The main patterns for the treatments in \figureref{fig:dp_notime} are as follows:

\begin{itemize}
\setlength\itemsep{0em}
    \item \textbf{Group 1. } Chemotherapy + Surgery + Hospitalization + Radiotherapy + Rehabilitation (11.3 \%)
    \item \textbf{Group 2. } Surgery + Hospitalization + Home hospitalization + Hormonotherapy (18.2 \%)
    \item \textbf{Group 3. } Surgery + Chemotherapy + Hospitalization + Radiotherapy (24\%)
    \item \textbf{Group 4. } Surgery + Radiotherapy + Hormonotherapy (5\%)
    \item \textbf{Group 5. } Surgery + Radiotherapy + Hormonotherapy (41.5\%)
\end{itemize}

\begin{figure}[h!]
\centering
\includegraphics[width=0.6\linewidth]{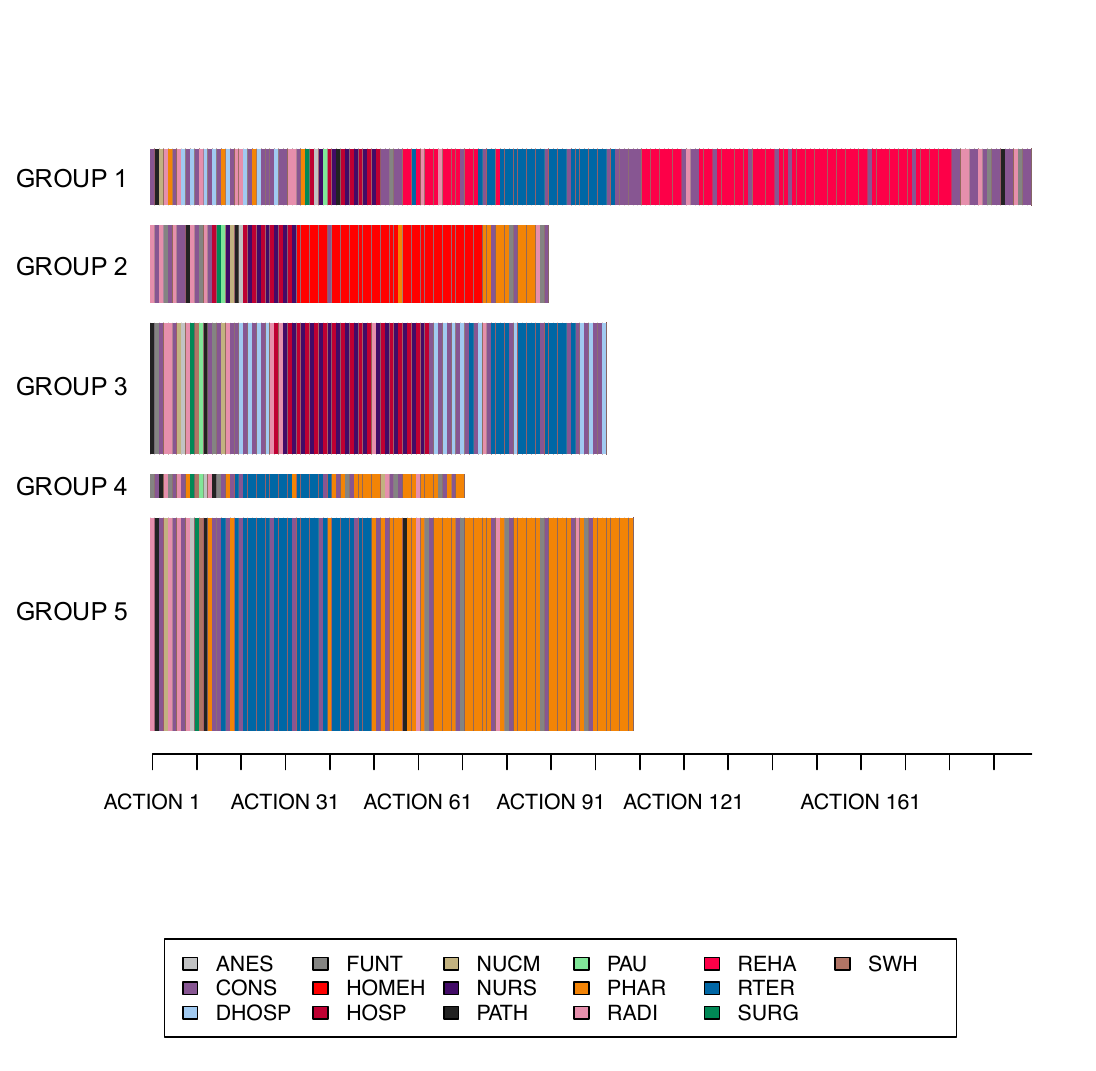}
\caption{Classification results for treatments associated with breast cancer without representing the time intervals between the medical actions.}
\label{app_fig:weibull_rep_notimeint}
\end{figure}

The major patterns of \figureref{app_fig:weibull_rep_notimeint} are as follows:
\begin{itemize}
\setlength\itemsep{0em}
    \item \textbf{Group 1. } Surgery + Chemotheray + Radiotherapy (25.7 \%)
    \item \textbf{Group 2. } Surgery + Radiotherapy (20.7 \%)
    \item \textbf{Group 3. } Surgery  + Hospitalization + Hormonotherapy (13.1\%)
    \item \textbf{Group 4. } Surgery + Radiotherapy + Hormonotherapy (23.3\%)
    \item \textbf{Group 5. } Chemotherapy + Surgery + Hospitalization + Radiotherapy + Chemotherapy (17.2\%)
\end{itemize}

Comparing the results of our time-dependent model with the representative sequences of actions obtained using the model developed in \citet{zaballa2023learning}, which does not consider the temporal component, we can identify several similarities. For instance, we can observe that the treatment patterns in Group 5 obtained from the time-dependent model (\figureref{app_fig:weibull_rep_notimeint}) match those in Group 4 obtained using the model in \citet{zaballa2023learning} (\figureref{fig:dp_notime}), although the proportion of patients assigned to these groups is different. Similarly, Group 2 from the time-dependent model (\figureref{app_fig:weibull_rep_notimeint}) and Group 3 from the model in \citet{zaballa2023learning} (\figureref{fig:dp_notime}) are also similar, with the exception that patients in Group 2 receive home hospitalization. However, there are slight variations in the remaining treatments between the two models.

\end{document}